%% file: main.tex
\documentclass[10pt,twocolumn,letterpaper]{article}

\usepackage{cvpr}              % To produce the CAMERA-READY version
\usepackage{graphicx}
\usepackage{amsmath}
\usepackage{amssymb}
\usepackage{booktabs}
\usepackage{mathtools}
\usepackage{multirow}
\usepackage{pifont}% http://ctan.org/pkg/pifont
\newcommand{\cmark}{\ding{51}}%
\newcommand{\xmark}{\ding{55}}%
\usepackage{fancyhdr}
\usepackage[ruled,vlined]{algorithm2e}

\usepackage{amsmath,amsfonts,amsthm,bm}
\newcommand{\method}{SLRNet}

\usepackage[pagebackref,breaklinks,colorlinks]{hyperref}
\usepackage{soul}

\usepackage[capitalize]{cleveref}
\crefname{section}{Sec.}{Secs.}
\Crefname{section}{Section}{Sections}
\Crefname{table}{Table}{Tables}
\crefname{table}{Tab.}{Tabs.}

\def\cvprPaperID{9305} % *** Enter the CVPR Paper ID here
\def\confName{CVPR}
\def\confYear{2022}

\begin{document}

\title{%\st{Similarity-based Label Reuse for Semi-Supervised Semantic Segmentation of Human Decomposition Images}
Semi-Supervised Semantic Segmentation Via Label Reuse for Human Decomposition Images}
%\title{Semi-Supervised Semantic Segmentation Via Label Reuse for Human Decomposition Images}
% zy: maybe change "Similarity" to another terms like "Deep Image Feature". 

\author{
%Sara Mousavi\\
%Department of Electrical Engineering and Computer Science\\
%The University of Tennessee Knoxville, USA\\
%{\tt\small mousavi@vols.utk.edu}
%\and
%Zhenning Yang\\
%Department of Electrical Engineering and Computer Science\\
%The University of Tennessee Knoxville, USA\\
%{\tt\small zyang37@vols.utk.edu}
%\and
%Kelly Cross\\
%Department of Anthropology\\
%The University of Tennessee Knoxville, USA\\
%{\tt\small kcross12@vols.utk.edu}
%\and
%Dawnie Steadman\\
%Department of Anthropology\\
%The University of Tennessee Knoxville, USA\\
%{\tt\small dsteadma@utk.edu }
%\and
%Audris Mockus\\
%Department of Electrical Engineering and Computer Science\\
%The University of Tennessee Knoxville, USA\\
%{\tt\small audris@utk.edu }
Sara Mousavi, Zhenning Yang, Kelly Cross, Dawnie Steadman, Audris Mockus \\
The University of Tennessee Knoxville, USA \\
%\{mousavi, zyang37, kcross12\}@vols.utk.edu \{dsteadma, audris\}@utk.edu
\{mousavi, zyang37, kcross12, dsteadma, audris\}@vols.utk.edu
}
\maketitle

\begin{abstract}

\input{abstract}
%   The ABSTRACT is to be in fully justified italicized text, at the top of the left-hand column, below the author and affiliation information.
%   Use the word ``Abstract'' as the title, in 12-point Times, boldface type, centered relative to the column, initially capitalized.
%   The abstract is to be in 10-point, single-spaced type.
%   Leave two blank lines after the Abstract, then begin the main text.
%   Look at previous CVPR abstracts to get a feel for style and length.
\end{abstract}

%%%%%%%%% BODY TEXT
\section{Introduction}
\label{sec:intro}
\input{intro}

%--------------------------
\section{Related Work}\label{sec:related_work}
\input{relatedwork}

%--------------------------
\section{Method}\label{sec:method}
\input{method}

%--------------------------
\section{Experimental Results}\label{sec:exp_results}
\input{results}

%---------------------------
\section{Conclusion}\label{sec:conclusion}
\input{conclusion}

\section{Supplementary Material}

In the main paper we have evaluated our method on an image dataset depicting decomposing human bodies. 
In this supplementary material, we first evaluate the generality of our proposed method for such datasets by conducting similar experiments on a dataset from a different domain capturing growing plants (Arabidopsis leaves from the Aberystwyth Leaf Evaluation Dataset~\cite{bell_jonathan_2016_168158}). Image data depicting different stages of growing plants manifest similarities to the human decomposition photos as both are depicting gradual changes over time that, over the full course of observation, lead to dramatic changes in appearance, but provide local similarities between neighboring timesteps that our method can leverage. Second, we provide more analysis on how a dataset such as the human decomposition data is different than a benchmark dataset such as VOC12. %Third, we provide results for applying a vanilla CycleGAN on the images of human decomposition as another comparison against a GAN-based method. 
Finally, we provide some edge cases for the pairs generated using our pairing algorithm from the human decomposition data. \textbf{CAUTION: This paper includes graphic content of human decomposition.}

%In the following sections, we first provide details about the Aberystwyth Leaf Evaluation dataset in Section~\ref{sec:leaf_data}, and then describe our experimental results on this dataset in Section~\ref{sec:leaf_results}.
%%%%%%%%%%%%%%%%%%%%%%%%%%%%%%%%%%%%%%%%%%%%%%%%%%%%%%%%%%%%%%%%%%%%%%%%%%%%%%%%%%%%%%%%%

\section{SLRNet for a Growing Plant Dataset}
In this section we examine the applicability of SLRNet for a similar dataset to the human decomposition dataset; Aberystwyth Leaf Evaluation dataset.  
\subsection{Aberystwyth Leaf Evaluation Dataset}\label{sec:leaf_data}
\begin{figure*}
    \centering
    \includegraphics[width=0.9\textwidth]{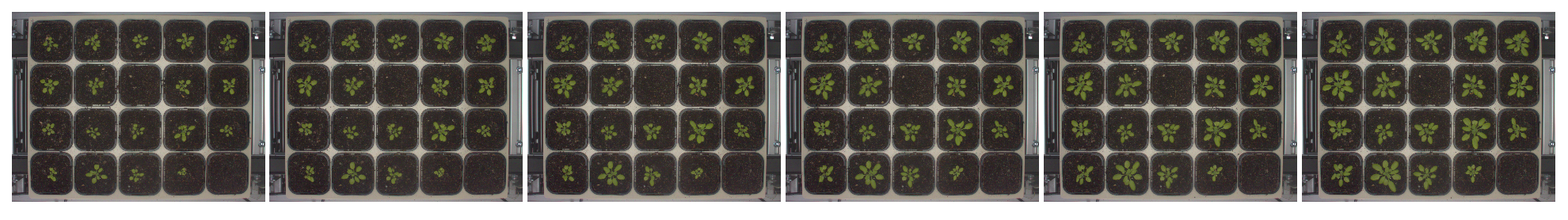}
    \caption{A set of example illustrating the gradual changes due to the growth.}
    \label{fig:plantexample}
\end{figure*}

 Aberystwyth Leaf Evaluation dataset is released by Aberystwyth University and has been collected to support researchers to further advance state-of-the-art methods used in image analysis for studying plants~\cite{bell_jonathan_2016_168158}. The images are collected by periodically taking pictures from Arabidopsis plants with 15-minute intervals using a robotic greenhouse system. The Aberystwyth dataset includes manual annotations for a subset of these image. The gradual growth of these plants results in a similar characteristic to the human decomposition data and provides an opportunity to utilize the limited available annotations and reuse them for the unlabeled images. An example showing a sequence of these images is shown in Figure \ref{fig:plantexample}.

\begin{figure*}
    \begin{center}
     \includegraphics[width=0.9\textwidth]{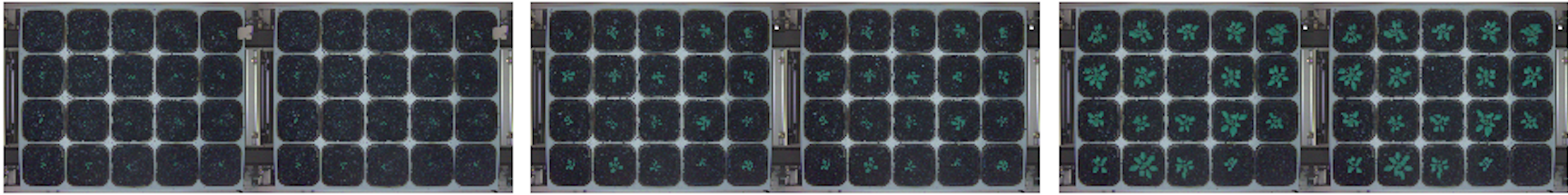}
    \caption{A few examples of pairing. Each labeled image is paired with an unlabeled image using our pairing algorithm.}
    \label{fig:plant_seq}   
    \end{center}
\end{figure*}
The Aberystwyth Leaf dataset records the growth of Arabidopsis Thaliana plants potted in four trays. This dataset includes $134,040$ Arabidopsis Thaliana plant images, from which $916$ are manually annotated. %Due to the gradual changes in the appearance of these plants, similar to the human decomposition data; they shared the characteristic which provides the opportunity for leveraging the large amount of unlabeled data for training. 

We use the same ratio as the human decomposition data, $60\%, 20\%, 20\%$, to create training, validation, and test sets respectively. Using our pairing algorithm, we use an additional $31,440$ individual Arabidopsis plant images paired with the labeled ones in the training process. A few paired examples are shown in Figure~\ref{fig:plant_seq}. In our experiments with this dataset, we set the number of classes to two, for ``background'' and ``leaf''.

%%%%%%%%%%%%%%%%%%%%%%%%%%%%%%%%%%%%%%%%%%%%%%%%%%%%%%%%%%%%%%%%%%%%%%%%%%%%%%%%%%%%%%%%%
\subsection{Experimental Results for Aberystwyth Leaf Evaluation Dataset}\label{sec:leaf_results}

We conducted two sets of experiments. First, to see the impact of the additional unlabeled data on the performance of our method in segmenting the Aberystwyth Leaf images, we compared the supervised version, when the model is only trained on the labeled data, vs. a semi-supervised version, when the model is trained on both labeled images and the pairs generated using our pairing algorithm. Second, we compared the performance of our method to that of two state-of-the-art methods, CCT~\cite{ouali2020semi} and PseudoSeg~\cite{zou2020pseudoseg} using the ResNet backbone for a fair comparison to CCT.
\begin{table}
    \caption{Supervised vs. semi-supervised comparisons of SLRNet with different backbones for segmenting images of Aberystwyth Leaf dataset. Results are shown in percentages.}
    \label{tbl:sup_semi}
    \begin{center}
    \begin{tabular}{ccccc}
    \hline
    \multicolumn{5}{c}{\textbf{SLRNet}}\\
    \hline
    \multirow{2}{*}{} & \multicolumn{2}{c}{\textbf{Supervised}} & \multicolumn{2}{c}{\textbf{Semi-supervised}} \\ \cline{2-5} 
     &\textbf{ mIoU} & \textbf{mAcc} & \textbf{mIoU} & \textbf{mAcc} \\ \hline
    \textbf{HRNet} & 92.43 & 97.28 & \textbf{94.27} & \textbf{98.03} \\ \hline
    \textbf{ResNet} & 88.17 & 95.64 & \textbf{94.2} & \textbf{98.01} \\ \hline
    \textbf{Xception} & 63.53 & 83.94 & \textbf{65.13} & \textbf{86.99} \\ \hline
    \end{tabular}
    \end{center}
\end{table}

\begin{table*}
    \caption{Mean-IoU and mean-pixel accuracy for CCT, PseudoSeg and SLRNet on the Aberystwyth Leaf test data using the ResNet backbone.}
    \label{tbl:plant_res}
    \begin{center}
    \begin{tabular}{lcccc}
    \hline
    \multirow{2}{*}{\textbf{}} & \multirow{2}{*}{\textbf{Mean-IoU (\%)}} & \multirow{2}{*}{\textbf{Mean-Acc (\%)}} & \multicolumn{2}{c}{\textbf{Per class IoU (\%)}} \\ \cline{4-5} 
                               &                                         &                                         & \textbf{BG}           & \textbf{Leaf}           \\ \hline
    \textbf{CCT}               & 51.73                                   & 82.59                                   & 81.71                 & 21.75                   \\ \hline
    \textbf{PseudoSeg}         & 90.64                                   & 96.7                                    & 95.92                 & 85.35                   \\ \hline
    % \textbf{SLRNet}            & \textbf{94.27}                          & \textbf{98.03}                          & 97.46                 & 90.80                   \\ \hline
    \textbf{SLRNet}            & \textbf{94.2}                          & \textbf{98.01}                          & \textbf{97.52}         & \textbf{90.88}                   \\ \hline
    \end{tabular}
    \end{center}
\end{table*}

\begin{figure*}
    \centering
    \includegraphics[width=0.9\textwidth]{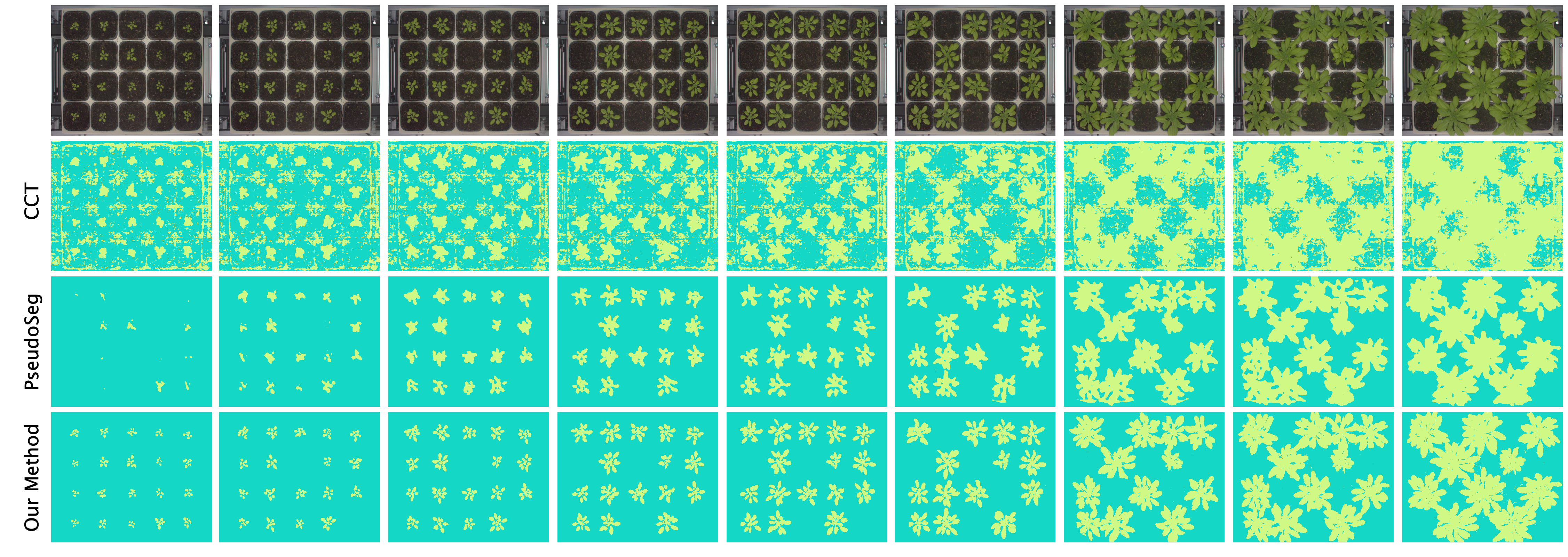}
    \caption{A few examples of performing semantic segmentation on the Aberystwyth Leaf dataset when using SLRNet, CCT and PseudoSeg.}
    \label{fig:plant_seg}
\end{figure*}

The result of the two experiments are shown in Tables~\ref{tbl:sup_semi} and~\ref{tbl:plant_res}. The results in Table~\ref{tbl:sup_semi} indicate that our method (SLRNet) indeed facilitates controlled use of the additional unlabeled data and the semi-supervised setting that includes this data performs noticeably better as compared to the supervised version. Furthermore, Table~\ref{tbl:plant_res} shows the results of comparing SLRNet with CCT and PseudoSeg.  The results indicate that our method consistently outperforms both CCT and PseudoSeg methods in all metrics. Notably, PseudoSeg and SLRNet perform much better on the plant dataset. This is not particularly surprising since only two classes are predicted (leaf and background) for plants in contrast to a much harder problem of $7$ classes for human decomposition. Additionally, CCT performed considerably worse than the other two methods in terms of its mean IoU. We suspect that CCT's complex structure and its various perturbations hinder its prediction of small leaves. Figure~\ref{fig:plant_seg} shows the predictions of these three methods for a few examples from Aberystwyth Leaf dataset. 
In summary, we found SLRNet to perform well on two unrelated photographic datasets, suggesting that it may be more generally applicable for datasets depicting subjects with gradual evolution. 

\section{Comparing Human Decomposition Images to Benchmark Datasets}
\begin{figure*}
    \centering
    \begin{subfigure}[b]{0.32\textwidth}
        \centering
        \includegraphics[width=\textwidth]{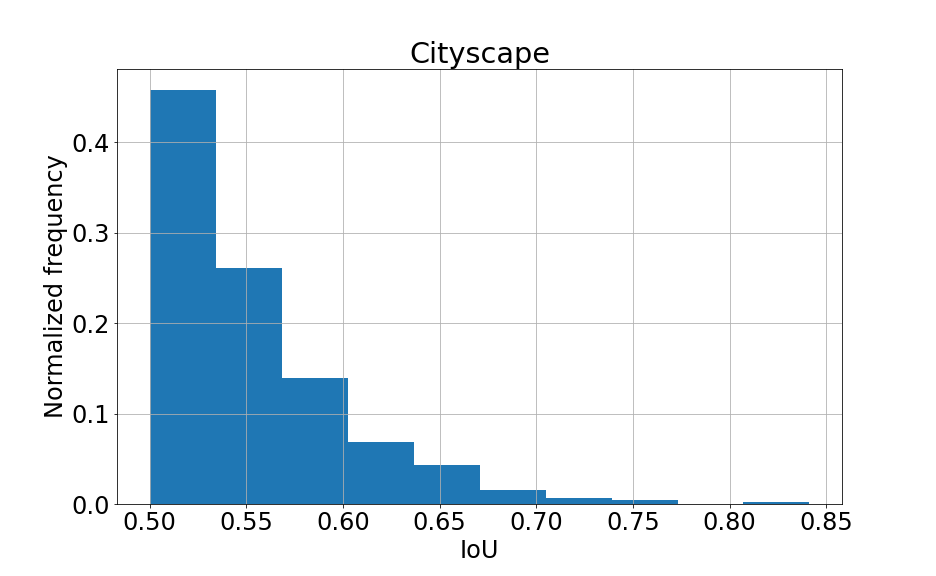}
        \caption[]%
        {{\small }}    
        \label{}
    \end{subfigure}
    \hfill
    \begin{subfigure}[b]{0.32\textwidth}  
    	\centering 
    	\includegraphics[width=\textwidth]{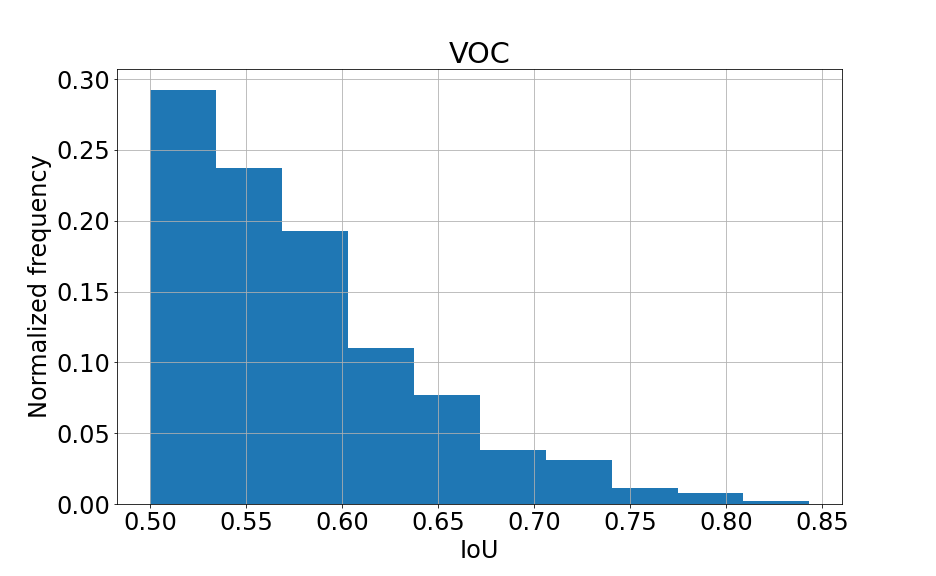}
    	\caption[]%
    	{{\small }}    
    	\label{}
    \end{subfigure}
    \begin{subfigure}[b]{0.32\textwidth}  
        \centering 
        \includegraphics[width=\textwidth]{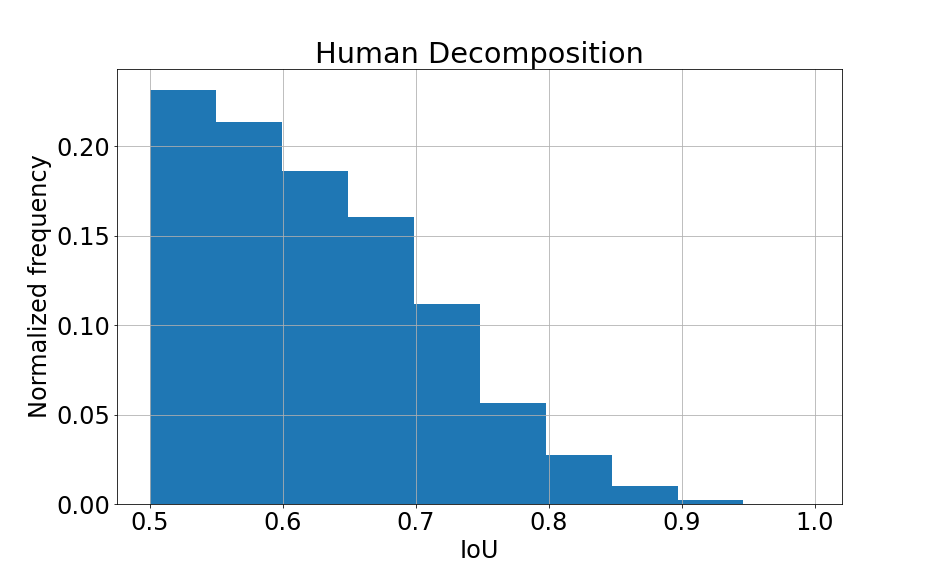}
        \caption[]%
        {{\small  }}    
        \label{}
    \end{subfigure}
    \caption{Histogram of IoU between pairs of labels for images with the same classes for Cityscape, Pascal VOC, and the human decomposition datasets are shown. The plots indicate that in a dataset with evolving content (gradual changes over time) such as the human decomposition dataset, there exist more label IoU similarity than others.}
    \label{fig:iou-hist}
\end{figure*}

SLRNet targets datasets representing evolving phenomena that have high potential for annotation reuse such as images of human decomposition, aging faces, growing plants, or decaying produce. Knowing the structure of such datasets and their gradual changes, such datasets are more likely to include images that can potentially have similar annotations (pixel-level labels). To confirm this hypothesis, we experimented on the Cityscape~\cite{Cordts_2016_CVPR}, VOC~\cite{everingham2007pascal}, and the human decomposition datasets. For each dataset, we first generated pairs of images depicting the exact same classes. We then calculated the intersection over union (IoU) between the labels of images in each pair to see how similar and align the labels of these images are. Normalized histograms of these calculated IoUs in Figure \ref{fig:iou-hist} show the result for this experiment. The result indicates that even though we have pairs of images with the same classes in all datasets, in the human decomposition dataset, these pairs tend to have similar masks with higher IoU than benchmark datasets VOC12 and Cityscape.

\section{Edge Cases of the Human Decomposition Pairs}
\begin{figure}
\centering
\begin{subfigure}{.5\textwidth}
  \centering
     \includegraphics[width=0.9\linewidth]{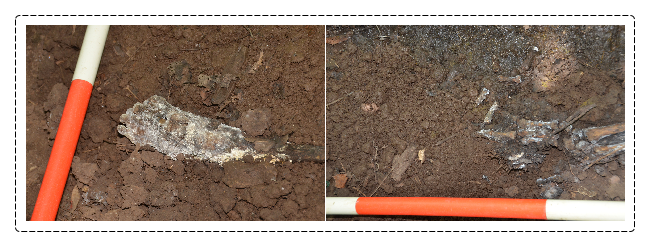}
    \caption{Example of a pair with different classes}
     \label{fig:badclass}
\end{subfigure}

\begin{subfigure}{.5\textwidth}
     \centering
     \includegraphics[width=0.9\linewidth]{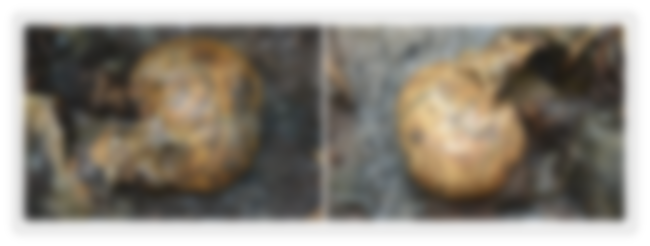}
    \caption{Example of an unaligned pair}
    \label{fig:badorientation}
\end{subfigure}
\caption{(a) shows examples of pairing images from different classes together. (b) shows examples of pairs with same classes but not perfectly aligned. Images are blurred due to their graphic nature.}
\label{fig:badpair}
\end{figure}

Based on the result provided in Section 4.4 in the main paper, our pairing algorithm is capable of pairing images from the same classes together ($95.06\%$ of the time). In this section we provide a few edge cases that can happen due to background-foreground similarity of the images. For example in Figure~\ref{fig:badclass}, a decayed hand and foot are paired together. In addition, Figure \ref{fig:badorientation} shows a pair in which images are not perfectly aligned but some part of the labels are still shared and reusable.

Manual observation of these pairs indicates that mismatch of classes rarely happens and that is when the background is a large portion of the image and the subjects in the two images are similar such as the example shown in Figure~\ref{fig:badclass}. Additionally, due to the characteristic of this dataset, even not perfectly aligned pairs as shown in Figure \ref{fig:badorientation} still provide correct a usable mask for some portion of the unlabeled images.

{\small
\bibliographystyle{ieee_fullname}
\bibliography{egbib}
}

\end{document}

%% file: abstract.tex
Semantic segmentation is a challenging computer vision task demanding a significant amount of pixel-level annotated data. 
Producing such data is a time-consuming and costly process, especially for domains with a scarcity of experts, such as medicine or forensic anthropology. While numerous semi-supervised approaches have been developed to make the most from the limited labeled data and ample amount of unlabeled data, domain-specific real-world datasets often have characteristics that both reduce the effectiveness of off-the-shelf state-of-the-art methods and also provide opportunities to create new methods that exploit these characteristics. We propose and evaluate a semi-supervised method that reuses available labels for unlabeled images of a dataset by exploiting existing similarities, while dynamically weighting the impact of these reused labels in the training process. %Specifically, we first use an unsupervised technique to find pairs of unlabeled and labeled images that are either directly or transitively most similar to one another. The pixel-level annotations of the labeled images are then reused as pseudo-labels for the unlabeled images in the pairs. Next, a segmentation network is trained using a custom loss function that dynamically controls the impact of the unlabeled images on the learning process depending on the level of similarity between the predicted annotations for the images in the pairs. 
We evaluate our method on a large dataset of human decomposition images and find that our method, while conceptually simple, outperforms state-of-the-art consistency and pseudo-labeling-based methods for the segmentation of this dataset. \textbf{CAUTION: This paper includes graphic content of human decomposition.}

%% file: intro.tex
%\textcolor{red}{Was starting with the second paragraph before (if the forensic paper was out you could quote it here :-)). Now we start with this: Semantic segmentation has many applications in various domains. In particular, segmenting body parts in images of human decomposition is a critical first step needed to make large and unique photographic datasets of human decomposition usable in forensic research. First, the type and level of decomposition varies among body parts~\cite{simmons2010debugging,galloway1989decay} and separating body parts with segmentation allows studying the presence of forensic features  on specific body parts (e.g. presence of mold on legs vs. other body parts). Second, it enables placing forensic features in context to pave the way for downstream models of time-of-death estimation that include forensic features and body parts as key predictors~\cite{gelderman2019estimation,canturk2018computational,boef2014sample}. Finally, it enables separating body parts from the background area and, therefore, helps with the identification of forensic features that can be very similar to the background (as shown in Figure \ref{fig:badexample}).}

\begin{figure}[b]
    \centering
    \includegraphics[width=0.9\columnwidth]{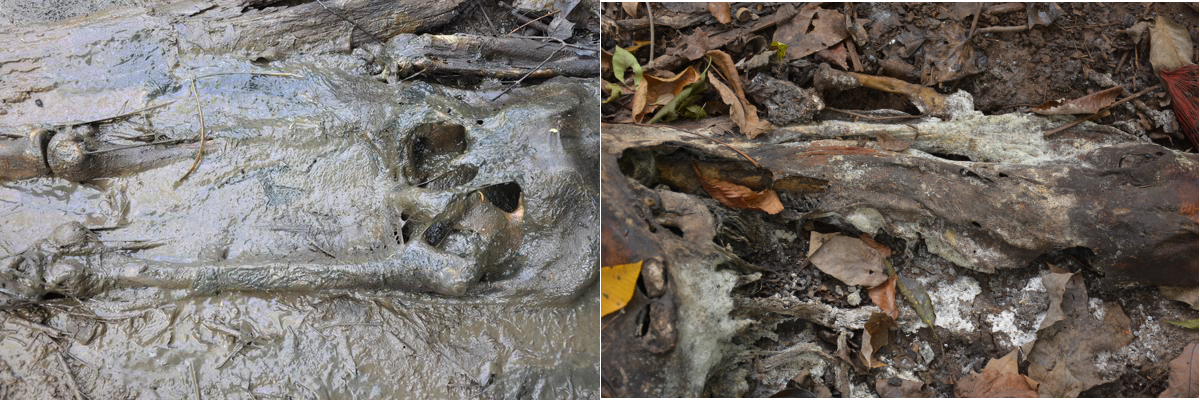}
    \caption{Examples of the background having a similar color and texture to the foreground in the human decomposition dataset.}
    \label{fig:badexample}
\end{figure}

Semantic segmentation has many applications in various domains. In particular, segmenting body parts in images of human decomposition is a critical first step needed to make large and unique photographic datasets of human decomposition usable in forensic research. First, the type and level of decomposition varies among body parts~\cite{simmons2010debugging,galloway1989decay} and separating body parts with segmentation allows studying the presence of forensic features  on specific body parts (e.g. presence of mold on legs vs. hands). Second, it enables placing forensic features in context to pave the way for downstream models of time-of-death estimation that include forensic features and body parts as key predictors~\cite{gelderman2019estimation,canturk2018computational,boef2014sample}. Finally, it enables separating body parts from the background area and, therefore, helps with the identification of forensic features that can be very similar to the background (as shown in Figure \ref{fig:badexample}).

Pixel-level annotations are necessary for accurate supervised semantic segmentation but may be too 
costly for many application domains. Manual pixel-level labeling needed for semantic segmentation can take $15$ to $60$ times longer than that of region-level and image-level labels~\cite{lin2014microsoft}. In addition, specialized expertise is often necessary and scarce for labeling images in domains such as medicine or forensic anthropology. To address such challenges, many semi-supervised and weakly-supervised methods have been developed to work with relatively few labeled and numerous unlabeled or weakly labeled images. 
%These methods are most often based on a few key assumptions of \textit{smoothness}: if $x$ and $x'$ are similar their labels $y$ and $y'$ should be the same, \textit{low density}: class boundaries should not pass through high-density regions, and \textit{manifold}: data points on the same low dimensional manifold should have the same label \cite{van2020survey}. %process of labeling is Although these limitations exist, some of these domains include \textcolor{red}{constraints/properties} in their data that can be leveraged for utilizing unlabeled data.

%%Semi-supervised methods have been studied extensively to leverage large amounts of unlabeled data. These methods are most often based on a few key assumptions of \textit{smoothness}: if $x$ and $x'$ are similar their labels $y$ and $y'$ should be the same, \textit{low density}: class boundaries should not pass through high density regions, and \textit{manifold}: data points on the same low dimensional manifold should have the same label \cite{van2020survey}.

%key ideas, namely, the use of pseudo-labels~\cite{wu2017semi}, consistency training~\cite{rasmus2015semi,miyato2018virtual}, bootstrapping~\cite{qiao2018deep}, and entropy minimization~\cite{grandvalet2005semi}.  \todo{additional details}

%\todo{fill out the citations}
Weakly-supervised segmentation methods~\cite{fang2018weakly,lee2019ficklenet} rely on weakly annotated data to produce annotations for the unlabeled portion of the data. The newly produced annotations along with the original existing ones are then used in a supervised manner for the task of interest. However, due to the need for weak labels in such methods, recently, semi-supervised segmentation methods have gained more attention. Many semi-supervised segmentation methods have been developed mainly based on 
%generating more data using GAN-based methods, 
producing pseudo-labels for existing unlabeled images \cite{lee2013pseudo,berthelot2019mixmatch,sohn2020fixmatch} or through consistency-based methods utilizing augmentations \cite{laine2018temporal,miyato2018virtual,tarvainen2017mean}, perturbations \cite{laine2018temporal}, and multi-model collaborations~\cite{ke2020guided}. 

While many of these methods have complex structures and their results highly depend on well-tailored perturbation techniques, in this work, we explore the possibility of exploiting intrinsic differences and similarities in the data itself. 
For example, in images depicting gradual growth or decay of subjects (plants or human bodies) from agriculture and anthropology domains, even though images depicting a specific instance of the same subject over time may look different, their annotation may be similar.

In this work, we tackle the problem of insufficient labels by presenting a semi-supervised method that exploits latent relationships among images. 
Our algorithm discovers such latent relationships in order to obtain pseudo-labels for unlabeled images by first pairing the labeled images with similar unlabeled images and then reusing available pixel-level annotations for similar but unlabeled images in each 
pair. We define a multi-objective loss function that penalizes the predictions depending on the level of predicted annotation similarity between the images in each pair. 
In other words, the key idea is to exploit the similarity present in the image collection by re-using annotations for unlabeled images weighted by the extent of the network's understanding of their label similarity while also jointly learning from supervised samples.
We call our method SLRNet for Similarity-based Label Reuse Network. The code is available at: [\textit{redacted for review}].%\footnote{Link to the repository: [redacted for review]}. %\href{https://github.com/saramsv/semantic-segmentation-pytorch}{https://github.com/saramsv/SLRNet}}.

%%While many large image datasets may contain groups of \textcolor{red}{images with similar annotation} %similar images
%%satisfying the key assumption of the proposed approach, we focus on evaluating our method on the human decomposition \textcolor{red}{dataset which is very important for forensic anthropology researchers.}

The human decomposition photographic collection used in this work is intended to support forensic research and casework, and contains over 1M photos taken from $500$ decaying subjects over dynamic intervals often days or weeks apart. At each timestep, one or more images depicting each class (body part) are present. %This dataset contains over 1M images of different body parts undergoing stages of decomposition over the span of approximately one year.

%Segmenting body parts in images of human decomposition is very important for forensic research for two reasons. First, it enables separating body parts from the background area and therefore, facilitates the identification of other forensic features. Second, the decomposition type varies among body parts~\cite{simmons2010debugging,galloway1989decay}, hence the same forensic feature needs to be placed in context to pave the way for downstream models of time-of-death estimation that include forensic features and body parts as key predictors~\cite{gelderman2019estimation,canturk2018computational,boef2014sample}.

Semantic segmentation of decaying body parts is complicated due to various reasons. First, the color and texture of decaying body parts and many forensic features such as mummified skin being extremely similar to the background (Figure \ref{fig:badexample}). Second, in human decomposition imagery, subjects gradually evolve over time from the ``fresh'' stage to the completely decayed stage of ``skeletonization''. Third, different body parts are difficult to distinguish due to environmental settings such as muddy ground, especially in late stages of decay. Finally, the often multi-day delay between photos, the lack of control in the camera view, changing weather conditions, disturbances caused by scavenging, and the lack of explicit links over time of classes limits the applicability of powerful video object segmentation (VOS) methods. However, the gradual decay of these subjects brings about a similarity attribute in such datasets that can be utilized towards developing an efficient but simple semi-supervised method.

In this work, we propose a simple method that reuses labeled data as pseudo-labels for unlabeled images. Specifically, we first use an unsupervised algorithm to identify immediately or transitively similar images to each labeled image. This algorithm exploits the fact that the images depicting the same evolving subjects are more similar if they are closer to each other on the decay spectrum, and the same subject may have a different appearance in early and late evolution stages, however have very similar annotations. We find similar images to a given labeled image among its neighbors recursively until no image with sufficiently high similarity is found. Pairs are then created between the annotated image and the similar images, with the annotation of the labeled images being applied to the unlabeled images as pseudo-labels. However, when training the segmentation network, we do not treat the losses from pseudo labels equally, but adjust them based on the level of similarity between the network's predictions for the images in the pairs. 

%In this work, we propose a simple but effective method that reuses human-provided labels as pseudo-labels for unlabeled images. We first use an unsupervised algorithm to identify immediately or transitively similar images to each labeled image. We then define a multi-objective loss function to jointly learn form the labeled images and their unlabeled matches by feeding them jointly to a segmentation network and reusing the existing label for the unlabeled ones to calculate the loss while dynamically weighting it. 

We evaluate our method on the human decomposition images and compare our method with two state-of-the-art semi-supervised semantic segmentation methods: CCT~\cite{ouali2020semi} and PseudoSeg~\cite{zou2020pseudoseg}. Results indicate that SLRNet, while having a much simpler conceptual structure and correspondingly shorter run-time, outperforms both CCT and PseudoSeg on evolving images of human decomposition.

In the rest of this paper, we start from related work in Section~\ref{sec:related_work} and the proposed method in Section~\ref{sec:method}. The dataset used in this work is presented in Section~\ref{sec:data} and the pairing algorithm in Section~\ref{sec:pair}. The training process is described in Section~\ref{sec:network} and our results and conclusion are presented in Sections~\ref{sec:exp_results} and~\ref{sec:conclusion}.

%% file: relatedwork.tex
\subsection{Semi-Supervised Learning}
Computer vision techniques are increasingly used to solve real-world problems in numerous domains, but 
the scarcity of clean labeled data hinders this progress. On one hand, supervised techniques need large training samples while unsupervised methods often do not achieve the desired accuracy. Semi-supervised methods \cite{weston2012deep,laine2017temporal,ganin2016domainadversarial,lin2015microsoft,kalluri2019universal} have emerged as a potential solution to this problem. They rely on a large set of unlabeled data and a limited labeled set and have recently achieved good performance. New efforts in this area take various approaches, such as the creation of pseudo-labels~\cite{lee2013pseudo,berthelot2019mixmatch,sohn2020fixmatch} or utilizing consistency-based techniques~\cite{miyato2018virtual,tarvainen2017mean,laine2017temporal} to counter over-fitting, increase the training sample, find the optimal boundary between classes, and improve generality of the models.

Semi-supervised methods are either transductive~\cite{liu2019deep,iscen2019label}, where the goal is to generate labels for a set of unlabeled images, or inductive ~\cite{wu2017semi,laine2018temporal,miyato2018virtual,tarvainen2017mean}, where the goal is to find a classifier that can predict labels for any image in the input space~\cite{van2020survey}. 

%such as 
%the creation of 
% creating
% pseudo-labels~\cite{lee2013pseudo,berthelot2019mixmatch,sohn2020fixmatch} or developing consistency-based approaches~\cite{miyato2018virtual,tarvainen2017mean,laine2017temporal} to counter over-fitting, increase the training sample, finding the optimal boundary between classes, and improving generality of the models.

% https://www.cs.sfu.ca/~anoop/papers/pdf/semisup_naacl.pdf
% where the goal is to not only generate labels for unlabeled images but also find a classifier that can predict labels for any images in the input space~\cite{van2020survey}. 

Transductive methods such as \cite{liu2019deep,iscen2019label} are based on label propagation to generate more training data. For example, Iscen et al. \cite{iscen2019label} use a transductive algorithm to generate pseudo-labels for unlabeled data. 
Liu et al. make use of labels from related domains and propagates them to generate pseudo-labels 
for a target dataset~\cite{liu2019deep}. Our work's pairing algorithm is also inspired by transductive methods in order to obtain pseudo-labels for unlabeled images (Section \ref{sec:pair}).

Some inductive-based methods use unlabeled data along with labeled data in a supervised fashion by first generating pseudo-labels for unlabeled images and then using a combination of both labeled and pseudo-labeled images to learn from%all labeled and unlabeled images together%and assigning pseudo-labels to the unlabeled ones
~\cite{wu2017semi}. Other methods enforce a type of consistency across predictions for unlabeled data \cite{laine2018temporal,miyato2018virtual,tarvainen2017mean}. Consistency-based methods rely on the idea that a network's predictions should be mostly similar across perturbations or slightly modified versions of the same input. For example, in the work by Laine et al. \cite{laine2018temporal}, consistency over augmented and dropped out versions of the input images is enforced. 
%where as Miyato et al. \cite{miyato2018virtual} use virtual adversarial training instead of random augmentation and dropout to maximize the effect of perturbations. 
Similar to consistency-based methods we consider images in a pair as perturbed versions of each other, however, unlike those methods we do not enforce consistency on the predictions of the network. Instead we use these predictions to control the impact of the pseudo-labeled images on the network's learning.

%------------------------------------------------------
\subsection{Semi-Supervised Semantic Segmentation}
In addition to image classification tasks, semi-supervised methods are also used for semantic segmentation.
Various methods \cite{souly2017semi,hung2018adversarial,ouali2020semi,zou2020pseudoseg,zhou2019collaborative} have been developed that achieve state-of-the-art on benchmark datasets such as VOC12 \cite{everingham2010pascal} and COCO \cite{lin2014microsoft} as well as domain specific data such as medical images \cite{decenciere2014feedback,porwal2018indian}.

%%Pseudo-label based methods such as \cite{fang2018weakly,lee2019ficklenet}, use weak labels to generate pseudo-labels for unlabeled images and then train a model in a semi-supervised fashion on both labeled and pseudo-labeled data. However, these weak labels are not available for many datasets.

%\todo{For comparison, the paper only compared two related works. The other stoa semi-supervised methods introduced in introduction and Sec.2.2 should be compared and discussed, such as GAN-based methods.}
 %\todo{explain why \cite{fang2018weakly} is not what we do. Sara: I looked at the paper. First of all this paper is a weakly supervised method when they have key-points available for all images and want to use the small portion of the data with body part annotation to generate body part annotation for the rest. There is two reasons why we can't use this method: 1) We don't have any key point data available, 2) we can't use a method to get key-points because our bodies are decayed and those methods perform poorly on them. }

% Generative Adversarial Networks (GANs) have been used in a semi-supervised learning setting~\cite{souly2017semi,hung2018adversarial} by generating adversarial examples using GANs for semantic segmentation and extending the generic GAN framework to pixel-level predictions.
% However, samples generated by these methods may not be sufficiently close to real images or labels to help the segmentation network; and other non-GAN-based methods have shown better performance ~\cite{ouali2020semi,zou2020pseudoseg}.

Consistency-based training has also been used for semantic segmentation with promising results. For example, CCT~\cite{ouali2020semi} is based on cross consistency learning and uses a segmentation network with an encoder and a decoder for annotated images. Additionally, CCT adds several auxiliary decoders that use perturbed versions of the encoder's output for the unlabeled data. It then enforces consistency over the outputs of the auxiliary decoders and that of the main decoder. CCT has achieved state-of-the-art results on the VOC12~\cite{everingham2010pascal} dataset.%, outperforming the aforementioned adversarial based methods~\cite{hung2018adversarial,souly2017semi}. %We, therefore, compare our proposed method to CCT. 

Other methods, such as PseudoSeg, combine ideas of pseudo-labeling and consistency learning~\cite{zou2020pseudoseg}. 
The authors use a similar idea to consistency-based methods to generate pseudo-labels. They fuse a self attention-based GradCAM of an unlabeled input image to their network's prediction for a weakly augmented version of that same input and use the result as a pseudo-label. Then, they enforce consistency between the predictions of the network for a strongly augmented version of that input and the pseudo-label. %\todo{maybe add more cites for cct and  pseduseg} 

%Our work is compared to CCT and PseudoSeg, for the purpose of semantic segmentation of human decomposition images. However, 

The performance of methods such as CCT and PseudoSeg highly depends on the types of perturbations and augmentations used in the process. In this work, we present a simple method that does not rely on external augmentation and perturbation, and is conceptually simpler. The key idea of our method is to find a way to effectively reuse the existing differences and similarities in the dataset itself.

%% file: method.tex
\begin{figure*}
    \centering
    \includegraphics[width=0.9\textwidth]{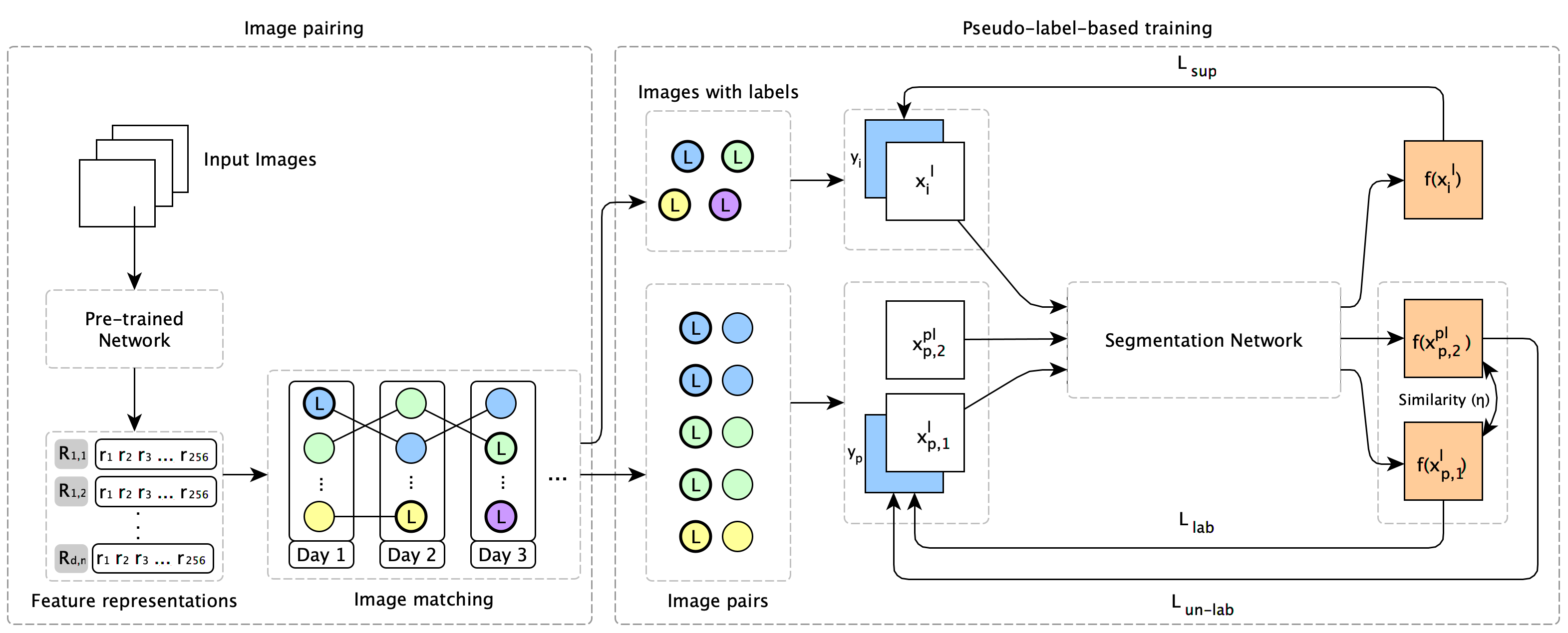}
    \caption{An overview of our method is shown (best seen in color). Images, first, go through a classification network which is pre-trained on ImageNet~\cite{deng2009imagenet} and fine-tuned on the human decomposition data to be mapped to feature vectors. Labeled images (shown with circles labeled with `L') and unlabeled images (circles with no labels) along with their feature vectors are then fed to our unsupervised image matching component to obtain similar images that can be paired with labeled ones and reuse the labels as pseudo-labels for the unlabeled images in the pairs. These pairs (one labeled image and one unlabeled image) along with the labeled images as a separate set, then go through our segmentation network that uses a customized loss function to control the effect of the pseudo-labels in the training process based on the similarity between the network's predictions ($\eta$) for the pair. 
    }
    \label{fig:overview}
\end{figure*}

Figure \ref{fig:overview} shows an overview of our method, which consists of two main steps: image-pairing and network training. In the pairing step, we use an unsupervised method %\todo{inspired by SChISM, the method cannot be explained that the difference}~\cite{mousavi2021schism} 
to identify unlabeled images that are similar to the labeled images, and therefore, can be paired with them. In the training phase, first, we simply reuse the annotation of the labeled images by assigning them as pseudo-labels to their similar but unlabeled match. We then use the image pairs along with the original labeled images as a separate set to train our semi-supervised semantic segmentation network in an end-to-end fashion. Our network is fed with both labeled images and the pairs at each iteration to jointly learn from both images with original labels and pseudo labels. Throughout the training, however, we use a custom loss function to control the effect of the unlabeled images and their pseudo-labels on the learning process with respect to the similarity of the two predictions for the images in the pairs as well as the current learning iteration.

In the following sections, we provide details on the dataset used in this work (Section \ref{sec:data}), the unsupervised image pairing process (Section \ref{sec:pair}), and the network structure and its training process (Section \ref{sec:network}).

%###########################################################################
% -dataset
\subsection{Human Decomposition Dataset}\label{sec:data}
%\subsubsection{Human Decomposition}\label{sec:hd}
The human decomposition dataset used in this work consists of photos taken from different body parts of various subjects donated to the forensic anthropology center of our university. The subjects were placed in an uncontrolled outdoor environment to record and track the decay process. The dataset has been collected over the years, with each subject staying in the facility for approximately one year. The photos are taken at various non-uniform intervals (one or more days apart) following a specific protocol. Each day, multiple images depicting various body parts for each subject are taken. The metadata associated with each image provides information about which donor the photo belongs to and the date at which the photo is taken. We denote these images by $\mathcal{I}_{d,n}$, where $d \in \{1, 2,\cdots,  D\}$, $n \in \{1, 2, \cdots, N\}$. $D$
is the total number of days photographed for each subject, $N$ is the total number of images taken at any given day, and $\mathcal{I}$ represents the set of all images for that subject. %The image resolutions vary from $2400 \times 1600$ up to $4900 \times 3200$. 
There exist $6$ main categories namely ``hand'', ``arm'',  ``foot'', ``leg'', ``torso'', and ``head''. In this work, we consider each body part as one class. %Note that typically there are one or more images representing the same area of the body in each day and each image may include multiple classes. Additionally, the same decomposition stage may correspond to multiple consecutive photography days and the number of days can vary for different body parts. For example, the first and last $3$ photography sessions might represent fresh and skeletal stages respectively. However, as a result of gradual decay and the limited number of total subjects, images from neighboring days tend to be more similar to one another than those from distant days.

%\subsubsection{Aberystwyth Leaf Evaluation}\label{sec:plant}
%Images depicting growing plants have similar characteristic to the human decomposition data in terms of gradual changes over time. Aberystwyth Leaf Evaluation dataset that is released by Aberystwyth University and has been collected to support researchers to further advance state-of-the-art methods used in image analysis for studying plants. The images are collected by periodically taking pictures from Arabidopsis plant with 15-minute intervals using a robotic greenhouse system. The Aberystwyth dataset includes manual annotations for a subset of these image~\cite{bell_jonathan_2016_168158}. 
%###########################################################################

\subsection{Unsupervised Image Pairing}\label{sec:pair}
\begin{figure}
    \centering
    \includegraphics[width=\columnwidth]{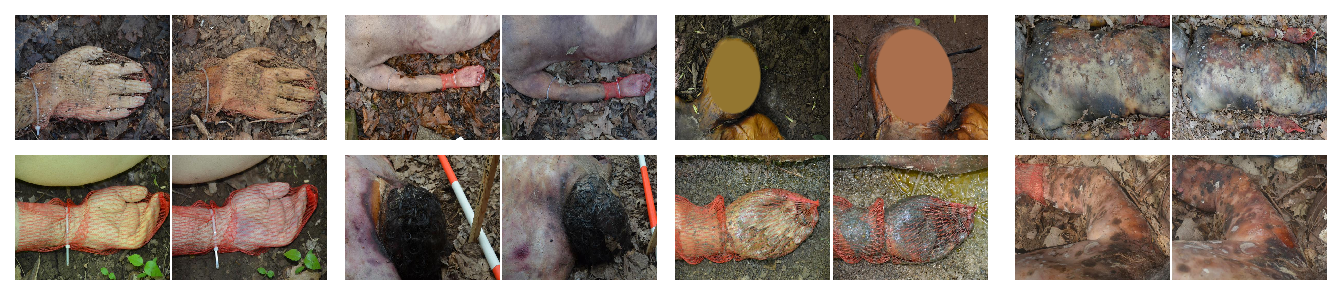}
    \caption{A few pair examples generated using our pairing algorithm. The third pair from the first row is anonymized}% described in Section \ref{sec:pair}}
    \label{fig:pairExample}
\end{figure}
The main idea of pairing labeled images with unlabeled ones is to utilize the potential reuse of annotation. It is likely for a dataset to have two or more images depicting the same content with only small differences. This is even more the case for image datasets with evolving content such as images of human decomposition, aging faces, growing plants, or decaying produce. 

In the human decomposition dataset, while photos from the same subjects at early evolution stages can drastically differ from those at late stages in terms of visual features, they still tend to have similar annotations. For example, while a hand in the ``fresh'' stage might look different from a hand in the late stages of decay, their annotations still resemble a hand from the same body and have similarities. Moreover, images belonging to neighboring days are more likely to be similar as well. %Inspired by SChISM~\cite{mousavi2021schism}, 
We leverage this characteristic and use neighbor-based comparison to identify the most suitable pairs for a given set of labeled images. 

First, we use the feature maps of a pre-trained classification network such as ResNet to map all images to their corresponding feature representations. To do so, we feed the images into the classification network, excluding its last fully-connected and softmax layers, and use the output vectors as the feature representations. 
In this work, we used ResNet50~\cite{he2016deep} pre-trained on ImageNet~\cite{deng2009imagenet} and fine-tuned with human decomposition data.
This fine-tuning uses only image-level labels and increases  the probability that similar images would contain the same body part. With ResNet, the vector length is $2048$. Inspired by DeepCluster~\cite{caron2018deep}, we also reduce the length of these representations to $256$ using Principle Component Analysis~\cite{wold1987principal} to improve the overall run-time of our method. Through experiments we found that on average, using PCA, pairing takes $54.42$\% less time than without PCA. We denote the resulting feature representation for $I_{d,n}$ which is the $n^{th}$ image in day $d$, as $R_{d,n}$. %In our experiments, using PCA for pairing a pool of $29,224$ images belonging to $5$ randomly selected subjects, on average, takes $54.42\%$ less time than without using PCA.} %On average, using PCA, the pairing process takes 54.42\% less time than without PCA

To identify potential images to pair with the labeled image $I_{d,n}^{l}$ (marked with an $l$ superscript), we compare its feature representation (i.e. $R_{d,n}^{l}$) with all other feature representations of its neighboring days and pick the most similar image as a match. From there, we recursively compare the matched images with their own respective neighbors to find other suitable pairs for the original $I_{d,n}^{l}$. %The number used as neighboring days for a given image is a hyper-parameter and is chosen through manual experiments. In this work, we use $4$ days.
%\st{The higher the neighboring days, the larger the pool of images to find matches from, and more matches are likely to be found while requiring more computation. Given this trade off, we chose $4$ for this hyper-parameter for our dataset.}

While this process can track similar images throughout different stages of decay, it also forces a match between each two neighboring days even if the difference is large. To circumvent this issue, we stop the matching process if the similarity of the two images being compared is lower than the overall average similarity calculated up to that point among the potential matches for $I_{d,n}^{l}$. Comparison of the image features is done through cosine similarity as: $Similarity(R_{d,n}, R_{d',m} ) = \frac{R_{d,n} . R_{d',m} }{\left \| R_{d,n} \right \| . \left \| R_{d',m} \right \| }$.
\begin{algorithm}[]
\caption{Image matching algorithm}
\label{alg:basic}
\footnotesize
\SetAlgoLined
      \KwResult{Pairs of labeled and unlabeled images}
      \SetKwFunction{FSub}{Compare}
      \SetKwFunction{FMain}{Main}
      \SetKwProg{Fn}{Def}{:}{}
      \Fn{\FMain{}}{
        matches = \{\}\;
        \For {each labeled image $I_{d,n}^{l}$}{
            Compare($I_{d,n}^{l}$, $I_{d,n}^{l}$)\;
        }
        pairs = [] \;
        \For {each labeled image $I_{d,n}^{l}$ in matches}{
            \For {each unlabeled match $I_{d,n}^{ul}$  in matches[$I_{d,n}^{l}$]}{
            pairs.append($I_{d,n}^{l}$, $I_{d,n}^{ul}$) \;
        }
        }
      }
      
      \Fn{\FSub{I, $ref$}}{
            R = PCA(Feature($ref$))\;
            \For {each neighboring day D for image $ref$} {
                maxSim = 0\;
                match = NULL\;
                \For {each image I' in D} {
                    R' = PCA(Feature(I'))\;
                    s = Similarity(R, R')\;
                    \If {$s > maxSim$} {
                        maxSim = s\;
                        match = I'\;
                    }
                }
                \If {$maxSim \ge AvgSim(matches[I])$} {
                    matches[I].add((match, maxSim))\;
                    Compare(I, match)\;
                }
            }
      }
\end{algorithm}
The pairing process is detailed in Alg. \ref{alg:basic} where, for each labeled image, initially the \textit{``Compare''} function is called for it and itself as \textit{``ref''}. In the \textit{``Compare''} function, each reference image (\textit{``ref''}) is compared to all of its neighboring images and those that are found to be appropriate matches will be added to the list of images that can be paired with the labeled image ($I_{d,n}^{l}$). The \textit{``Compare''} function is then recursively called for each image in the match set. 

From the matches obtained for $I_{d,n}^{l}$, a few may coincidentally have labels. Those are removed from the group of images being paired with $I_{d,n}^{l}$ as they will be later paired with unlabeled matches of their own. A few example pairs are shown in Figure \ref{fig:pairExample}. %and \ref{fig:plant_pair_example}.

% \begin{figure}
%         \centering
%         \includegraphics[width=\columnwidth]{imgs/plants_seqs3.png}
%         \caption{An example illustrating evolution in growing plants in the Aberystwyth dataset. }
%         \label{fig:plant_pair_example}
%     \end{figure}
%###########################################################################
\subsection{Network Structure and Training}\label{sec:network}
%\todo{the segmentation network consists of an encoder, a decoder and a custom loss function. What’s the structure of encoder and decoder? }
%\todo{(reviewer #4) The HRNet used in the experiment maintains a high resolution output, so Decoder is not needed, which is not consistent with Table 3 ???}

we denote the set of labeled images by
  $\mathcal{I}^l=\left \{ ( x_{1}^{l},y_{1}  ),( x_{2}^{l},y_{2} ),\cdots  ( x_{L}^{l},y_{L} ) \right \}$,
where $L$ is the total number of labeled images, $y_{i}$ is the annotation for the $i^{th}$ labeled image (i.e. $x_{i}^{l}$) and has dimensions of $W \times H \times C$ representing width, height and the number of classes, respectively. 

We reuse the annotations of the labeled images by assigning them as pseudo-labels (marked with a $pl$ superscript to denote that they have pseudo-labels) to the image with no labels in the pairs obtained from our image pairing algorithm described in the previous section (\ref{sec:pair}). We denote the pairs with: $\mathcal{P}=\left\{( (x_{1,1}^{l},x_{1,2}^{pl}), y_{1}^{}),( (x_{2,1}^{l},x_{2,2}^{pl}) ,y_{2}),\cdots( \\
(x_{P,1}^{l},x_{P,2}^{pl}) ,y_{P}^{}) \right\}$. where $P$ is the total number of pairs and $x_{p,1}^{l}$ and $x_{p,2}^{pl}$ are the first labeled and second pseudo-labeled elements in the $p$th pair respectively. In this setting $\left \{  x_{1,1}^{l}, x_{2,1}^{l} ,\cdots, x_{P,1}^{l} \right \} \subset \mathcal{I}^l$ and these are not unique images, whereas $\left \{ x_{1,2}^{pl} , x_{2,2}^{pl} ,\cdots , x_{P,2}^{pl}  \right \}$ are $P$ unique unlabeled images. That is because a labeled image could be paired with more than one image. 

It is important to note that there may not exist an unlabeled match for every single image in the labeled set. Therefore, we feed the labeled images in a separate set, parallel to the pairs, even though some of them might be already included in the pair branch, to ensure the inclusion of all labeled images in the training process. 

Additionally, as is expected, the location and orientation of the body parts captured in an unlabeled image may not be perfectly aligned with that of the labeled image in the pair even though the photos are taken following a specific protocol. Therefore, we introduce a custom loss function to control the impact of the pseudo-labels on the learning process with respect to the level of similarity between the network's predicted labels for the two images in the pair.

Our objective is to exploit the additional pseudo-labeled images to improve the performance of the semantic segmentation network. In this method, 
%the segmentation network consists of an encoder, a decoder, and as previously mentioned, 
we use a semantic segmentation network with a multi-objective loss function to facilitate learning from both labeled and pseudo-labeled images. At each iteration, the network is fed with a labeled image and a pair of images. We calculate a supervised loss and a loss for pairs, following Eq. \ref{equ:sup_loss} and Eq. \ref{equ:pair_loss} respectively. In the supervised loss calculated for the labeled images (Eq. \ref{equ:sup_loss}), $CE$ is cross entropy calculated using the ground truth $y_i$ and the predicted labels for input $x_i^{l}$ which is shown by $y_i^{'} = f(x_i^{l})$. 

\begin{equation}\label{equ:sup_loss}
\mathcal{L}_{sup} = \frac{1}{\left|\mathcal{I}^l\right|}\sum_{x_i^{l},y_i \in \mathcal{I}^l}CE (y_i, y_i^{'})
\end{equation}

Next, we calculate a loss for the pairs. As mentioned above, each pair has one labeled and one unlabeled image. We calculate pair loss following Eq. \ref{equ:pair_loss}. 
\begin{equation}\label{equ:pair_loss}
\begin{multlined}
\mathcal{L}_{pair} = \mathcal{L}_{lab} + \mathcal{L}_{un-lab}= \frac{1}{\left|\mathcal{P}\right|}\sum_{x_{1,i}^{l},y_i \in \mathcal{P}} CE \big(y_i, f(x_{1,i}^{l})\big)\\ + 
\frac{1}{\left|\mathcal{P}\right|}\sum_{x_{2,i}^{pl},y_i \in \mathcal{P}} \lambda * \eta * CE \big(y_i, f(x_{2,i}^{pl})\big) 
\end{multlined}
\end{equation}

\noindent where $ i\in \{1,2,...,P\}$. The first and the second part of the loss are calculated based on the labeled image, $y_{i}$ and $f(x_{1,i}^{l})$, and the unlabeled image, $y_{i}$ and $f(x_{2,i}^{pl})$, in the pair respectively. Since $y_{i}$ are not actual labels for $x_{2,i}^{pl}$, the loss calculated based on them is weighted by a similarity-based weight, $\eta$, and an iteration-based weight, $\lambda$. We use the prediction of the network for the pair, to determine the level of contribution for the loss calculated based on the pseudo labels in the backpropagation and calculate $\eta$ accordingly. The idea is to use the network's understanding to measure how similar and aligned the annotations for the images in a pair are. We compare the prediction of the network for the images in the pair to each other. The higher the similarity, the larger the weight will be. If the prediction of the network for the two images is different, the images may not be well aligned and the weight for the loss calculated based on the pseudo-label will be smaller. $\eta$ is at most $1$. %The value of $\eta$ is at most $1$ and calculated as:
\begin{equation}\label{equ:eta}
\eta = \frac{f(x_{1,i}^{l}) \cap f(x_{2,i}^{pl})}{(W \times H)_{x_{1,i}^{l}}}
\end{equation}
\noindent where $W$ and $H$ are the width and the height for $x_{1,i}^{l}$ respectively. Furthermore, $f(x_{1,i}^{l})$ and $f(x_{2,i}^{pl})$ are the predictions of the network for $x_{1,i}^{l}$ and $x_{2,i}^{pl}$ respectively.
%If these annotations are not similar or aligned enough, we reduce the weight of the loss calculated based on the pseudo-label and penalize the network mainly based on the supervised loss

Additionally, since the network's prediction is not robust at early epochs, we use $\lambda$ calculated based on the iteration numbers to minimize the influence of the initial noisy predictions and incorrect use of the pseudo labels in the training process. The value of $\lambda$ linearly increases with training iteration and is at most $1$.
\begin{equation}\label{equ:lambda}
\lambda = \frac{(epoch * ipe + iter)}{max\_iters}
\end{equation}
 where $epoch$ is the current epoch, $ipe$ is the number of iterations per each epoch, $iter$ is the current iteration, and $max\_iter$ is $the\ total\_number\_of\_epochs \times ipe$. The network is trained to minimize the overall loss:% that is calculated as $\mathcal{L} = \mathcal{L}_{sup} + \mathcal{L}_{pair}$.
\begin{equation}\label{equ:total_loss}
\mathcal{L} = \mathcal{L}_{sup} + \mathcal{L}_{pair}
\end{equation}

%% file: results.tex
%In this section, we provide implementation details and evaluation of our method.
In this section, we provide implementation details and an evaluation of our method. 
%(Section \ref{sec:implementationdetail}), train and test data as well as the evaluation metrics (\ref{sec:data&metrc}), comparison to state-of-art methods (\ref{sec:comparisons}), evaluation of our pairing algorithm (\ref{sec:evaluation}), and an ablation study on our network's components (\ref{sec:Ablationstudy}). 

%################################# tables ##########################################
\begin{table*}[]
\small
\caption{Mean-IoU, mean-pixel accuracy, per class IoU, and run-time for CCT, PseudoSeg and~\method~on the test data. The results indicate that our method consistently outperforms other methods on most classes with a large margin. Results are in percentages.}% HRNet backbone is better capable of segmenting small classes such as ``hand'' and ``foot'' as well as consistently performing well on most classes.}
\begin{center}
\label{tbl:compare-to-cct}
\begin{tabular}{llcccccccccl}
\hline
\multirow{2}{*}{\textbf{}} & \multirow{2}{*}{\textbf{Backbone}} & \multirow{2}{*}{\textbf{mIoU}} & \multirow{2}{*}{\textbf{mAcc}} & \multicolumn{7}{c}{\textbf{Per Class IoU}} & \multirow{2}{*}{\textbf{Run time}} 
\\ \cline{5-11}
 &  &  &  & \textbf{BG} & \textbf{Foot} & \textbf{Hand} & \textbf{Arm} & \textbf{Leg} & \textbf{Torso} & \textbf{Head} &   \\ \hline
\textbf{CCT} & ResNet & 62.77 & 86.67 & 88.73 & 52.16 & 53.29 & 48.56 & 63.39 & 60.67 & 72.57 & 23967m26s \\ \hline
\multirow{2}{*}{\textbf{PseudoSeg}} & Xception & 62.22 & 86.62 & 88.45 & 48.47 &  49.11 & 43.88 & 61.51 & 63.93 & 80.16 & 1334m5s \\
%Xception & 65.81 & 88.28 & 89.24 & 53.96 & 50.51 & 52.37 & 66.72 & 68.36 & 79.52 & 819m14.057s \\
% & ResNet & 65.15 & 87.82 & 88.91 & 50.37 & 53.35 & 52.87 & 65.03 & 65.76 & 79.79  \\ \hline
 & ResNet & 62.49 & 86.99 & 88.56 & 45.84 & 47.16 & 50.46 & 62.74 & 64.04 & 78.59 & 1500m17s \\ \hline
\multirow{3}{*}{\textbf{\begin{tabular}[c]{@{}c@{}}Our\\ Method\end{tabular}}} & Xception & 66.76 & 88.27 & 88.63 & 78.63 & 78.63 & 78.63 & 78.63 & 78.63 & 78.63 & 1300m19s \\
 & ResNet & 65.3 & 87.12 & 87.44 & 57.28 & 58.07 & 49.68 & 64.32 & 64.75 & 75.59 & 764m15s \\
 %& HRNet & \textbf{71.55} & \textbf{89.77} & \textbf{90.30} & \textbf{60.79} & \textbf{67.37} & \textbf{60.57} & \textbf{68.13} & \textbf{70.99} & \textbf{82.69} & 708m24.637s \\ \hline
 & HRNet & \textbf{72.42} & \textbf{90.04} & \textbf{90.01} & \textbf{62.42} & \textbf{67.9} & \textbf{61.06} & \textbf{69.5} & \textbf{72.06} & \textbf{83.98} & 708m24s \\ \hline
\end{tabular}
\end{center}
\end{table*}

%################################# table ##########################################
% \begin{table}[]
% \begin{tabular}{ccccc}
% \hline
% \multirow{2}{*}{} & \multicolumn{2}{c}{\textbf{Supervised}} & \multicolumn{2}{c}{\textbf{Semi-supervised}} \\ \cline{2-5} 
%  & IoU (\%) & Acc (\%) & IoU (\%) & Acc (\%) \\ \hline
% HRNet & 92.43 & 97.28 & \textbf{94.27} & \textbf{98.03} \\ \hline
% ResNet & 88.17 & 95.64 & \textbf{94.2} & \textbf{98.01} \\ \hline
% Xception & 63.53 & 83.94 & \textbf{65.13} & \textbf{86.99} \\ \hline
% \end{tabular}
% \end{table}

%################################# table ##########################################

\begin{table*}[]
\small
\caption{Ablation study to examine the effect of $\lambda$, $\eta$, the additional unlabeled data produced using our pairing algorithm, and the quality of the pairs on the overall performance of our method with different backbones. Results are in percentages.}
\label{tbl:ablation}
\begin{center}
\begin{tabular}{lcccccccccc}
\hline
\multicolumn{1}{l}{}            & \multicolumn{4}{c}{}                                                              & \multicolumn{2}{c}{\textbf{HRNet}} & \multicolumn{2}{c}{\textbf{ResNet}} & \multicolumn{2}{c}{\textbf{Xception}} \\ \hline
\multicolumn{1}{l}{}            & \begin{tabular}[c]{@{}c@{}}\textbf{Labeled}\\ \textbf{Images}\end{tabular} & \textbf{Pairs} & \bm{$\lambda$} & \bm{$\eta$}   & \textbf{mIoU}         & \textbf{mAcc}        & \textbf{mIoU}           & \textbf{mAcc}           & \textbf{mIoU}          & \textbf{mAcc}           \\ \hline

\textbf{Supervised}                  & \cmark  & \xmark & \xmark  & \xmark & 68.42            & 88.86           & 47.38             & 80.11              & 61.15             & 86.39             \\ \hline
\multirow{3}{*}{\textbf{Semi-supervised}} &   \cmark                                                    & \cmark & \xmark  & \xmark & 61.37            & 85.94           & 54.53             & 82.04              & 63.46             & 86.44       \\
                                & \cmark                                                    & \cmark & \xmark  & \cmark & 63.97            & 86.24           & 60.71             & 85.05              & 63.65             & \textbf{88.71}             \\
                                & \cmark                                                    & \cmark & \cmark  & \xmark & 67.18            & 88.37           &           59.89        &     84.53               &     62.24              &      86.41             \\ 
                                & \cmark                                                    & \cmark & \cmark  & \cmark & \textbf{72.42}   & \textbf{90.04}  & \textbf{65.3}     & \textbf{87.12}     & \textbf{66.76}    & 88.27        \\\hline
                                \textbf{Semi-supervised}  & \cmark     & Random & \cmark  & \cmark & 69.43   & 89.38   & 58.53  & 85.81   & 62.85 & 86.84    \\ \hline

\end{tabular}
\end{center}
\end{table*}

%\todo{Furthermore, we evaluate our method on Aberystwyth plant dataset. The result is shown in Table \ref{tbl:plant}.}
%#########################Implementation Details####################################################
\subsection{Implementation Details}\label{sec:implementationdetail}
We implemented our method using %three different segmentation networks: 
HRNetV2\cite{sun2019high}, Xception~\cite{chollet2017xception} and ResNet~\cite{he2016deep} as the backbones built on the MIT implementation \cite{semantic-segmentation-pytorch}.
We implemented SLRNet using the PyTorch framework \cite{NEURIPS2019_9015}. We trained our method on a single $Tesla V100-SXM2$ GPU with $32GB$ memory. To train the segmentation network, we used Stochastic Gradient Descent (SGD) \cite{ketkar2017stochastic} with momentum of $0.9$ and weight decay of $10^{-4}$. We started with a learning rate of $0.02$. The learning rate is gradually decreased using polynomial decay with power $0.9$ \cite{chen2017deeplab}. We used the number of train samples divided by batch size as iterations per epoch. 

%########################### Dataset and Evaluation Metric #######################################
\subsection{Dataset and Evaluation Metric}\label{sec:data&metrc}
%In this work, we use a  human decomposition dataset with over one million images that are collected through daily photos of 500 subjects. 
%and a plant dataset with a total of 6702 images (more detail is provided in Section \ref{sec:data}). 

The human decomposition dataset includes $1864$ annotated images for ``hand'', ``arm'', ``foot'', ``leg'', ``torso'', and ``head'' classes. We use $60\%, 20\%, 20\%$ ratio to create training, validation and test sets. As a result, we have $1118$, $373$, and $373$ images in our training, validation and test sets respectively. Additionally, we use $5906$ unlabeled images resulted from identifying matches for the labeled images in the training set with other unlabeled images in the dataset using our pairing algorithm. 
Similar to other semantic segmentation works \cite{hung2018adversarial,ouali2020semi,souly2017semi}, we use the mean intersection-over-union (mean IoU) and pixel accuracy as evaluation metrics.
We also provide additional evaluation for a plant dataset (that shares similar characteristics to the human decomposition dataset) in the supplementary material.

%The Aberystwyth Leaf dataset records the growth of Arabidopsis Thaliana plants potted in four trays. This dataset includes $134040$ Arabidopsis Thaliana plants, from which $916$ are manually annotated. Due to the gradual changes in the appearance of these plants, similar to the human decomposition data; they shared the characteristic which provides the opportunity for leveraging the large amount of unlabeled data for training. 
%We use the same ratio as the human decomposition data ( $60\%, 20\%, 20\%$) to create training, validation, and test sets. Using our pairing algorithm, we use additional $31440$ individual Arabidopsis plants paired with the labeled ones in the training process.

%##################### Comparisons to Previous Work ######################################
\subsection{Comparisons to Previous Work}\label{sec:comparisons}
To the best of our knowledge, we are the first to explore semantic segmentation for images with evolving content such as human decomposition data and therefore there are no similar benchmarks or state-of-the-art methods on this topic. Therefore, to evaluate the effectiveness of our method, we quantitatively compare it with previous general state-of-the-art semi-supervised semantic segmentation methods, namely %on PASCAL VOC dataset \cite{everingham2010pascal}, called
CCT~\cite{ouali2020semi} and PseudoSeg~\cite{zou2020pseudoseg}. 

CCT is a consistency-based method that enforces consistency to the network's predictions for various perturbed version of an input. It uses a  two branch training structure one for labeled and one for unlabeled data. The two branches share the same encoder and one decoder. On the unsupervised branch it uses $K=7$ auxiliary decoders and various perturbations %(minimum of $2$ and maximum of $6$) 
and enforces consistency between their outputs and the output from the main shared decoder on the same input through a loss function. 

PseudoSeg uses a mix of pseudo-labeling and consistency-based training to leverage unlabeled images in the network's learning~\cite{zou2020pseudoseg}. In PseudoSeg, the pseudo-labels are generated by fusing the network prediction for a weakly augmented input image and the self-attention GradCAM generated for that input. In the training process, the authors use a similar idea to consistency-based methods and force their network's prediction for a strongly augmented version of the same input to be consistent with the pseudo-label that resulted from the fusion process.

To compare our method to CCT and PseudoSeg, we applied their semi-supervised setting to our data and compared the results with those obtained using our method. CCT and PseudoSeg use both labeled and unlabeled images to learn from them jointly. We use the unlabeled images obtained from our pairing algorithm and the labeled images as their unlabeled and labeled training inputs, respectively. We use the same validation and test sets for our method, CCT, and PseudoSeg. The results of comparing our method to CCT and PseudoSeg are shown Table~\ref{tbl:compare-to-cct}. 

%\todo{Compared to PseudoSeg, the comparison experimental results using ResNet backbone are relatively poor in some metrics (e.g., pixel acc, arm, leg, torso and head per class IoU). What is the reason? I didn't see related discussions in the paper. So, it is insufficient to reflect the strengths of the proposed method}

The results indicate that our method outperforms both CCT and PseudoSeg using the mean-IoU and mean-Acc metrics with a few minor exceptions for ``BG'', ``Arm'', and ``Head'' classes in PseudoSeg. %Examining the predictions, it appears that PseudoSeg predicts more false positives for these classes (one such example is shown in the last row in Figure \ref{fig:iccv_result} where some parts of the ``foot'' is predicted as ``head''). 
The results also confirm that, in general, the HRNet backbone captures spatial information better and performs well even for classes that occupy fewer pixels than the others and outperforms the other backbones for all methods. In the human decomposition data used in this work, this difference is pronounced due to the need to maintain multiple feature resolutions since we have images of the same class with varying views. Furthermore, the run-time reported in Table \ref{tbl:compare-to-cct} shows that SLRNet has a shorter run time than CCT and PseudoSeg while having a conceptually simpler structure.

%Our examination of the predictions done by PseudoSeg and SLRNet on the test set show that there are more false positive predictions for these classes in the PseudoSeg predictions and more false negatives for these classes in SLRNet predictions with ResNet backbone.  which increase the chance of correctly 

%\todo{PseudoSeg is not worse than the method proposed in this paper when using the same backbone, resnet, how should the authors explain this?}
%In some classes with a clear shape that decay does not destroy the structure of the class, like ``head'', PseudoSeg with ResNet backbone tents to perform better. However,
%################################# Ablation Study ##########################################
\subsection{Image Pairing Evaluation }\label{sec:evaluation}

To evaluate the pairing component of SLRNet, we first analyzed the effect of the quality of the pairs on the performance by replacing them with random ones where each labeled image is paired with a random unlabeled image. The results in Table \ref{tbl:ablation} show that using random pairs reduces the performance of the model as expected. 

We further evaluated the pairing algorithm by examining to what degree the generated pairs represent the same classes. To do so, we selected the top $5$ subjects with the highest number of labeled images, 
%with a total of $29,224$ images
for which $107$ images were pixel-level labeled. Applying our pairing algorithm to this data resulted in $1343$ pairs. To perform this evaluation, we manually image-level labeled unlabeled images in the pairs. This experiment showed that in $95.06\%$ of the pairs, both images represented the same classes. 
We also evaluated our pairing algorithm's quality by comparing the IoU of the labels in pairs generated using our algorithm with the mean and median IoUs of all possible pairs from same-class images in our labeled data, chosen as a baseline.
%vs. exhaustively pairing images with the same classes together for all labeled images of a given subject. 
Our pairs resulted in mean and median IoUs of $47.36\%$ and $41.15\%$ while pairing all same-class images resulted in $24.7\%$ and $19.18\%$ mean and median IoUs, respectively, indicating that our method was able to select higher quality matches than the average quality of matches in same-class pairs.

% pairing:
% count  48.000000
% mean    0.473634
% std     0.221444
% min     0.027549
% 25%     0.292106
% 50%     0.411575
% 75%     0.679515
% max     0.798334

% class-based-pairing
% count    886.000000
% mean       0.246973
% std        0.205054
% min        0.000000
% 25%        0.080769
% 50%        0.191865
% 75%        0.382387
% max        0.920375

\subsection{Ablation Study}\label{sec:Ablationstudy}
%We use an ablation study shown in Table \ref{tbl:ablation} to examine the impact of $\lambda$, $\eta$, and the additional unlabeled data produced using our pairing algorithm on the overall performance of our model. 
SLRNet, as a whole, uses both labeled and paired images, weights the loss calculated based on the pseudo-labels using $\eta$, and reduces the initial noisy and unstable behavior of the network using $\lambda$. We conduct an ablation study by assessing the performance of our method in the absence of these components. The scenarios shown in Table \ref{tbl:ablation} are 1) the presence of labeled images, pairs, $\lambda$, and $\eta$, 2) the presence of labeled images, pairs, and $\eta$ and the absence of $\lambda$, 3) the presence of labeled images, pairs, and $\lambda$ and the absence of $\eta$, 4) the presence of labeled images, pairs and the absence of both $\lambda$ and $\eta$, 5) the presence of labeled images and the absence of pairs, $\lambda$, and $\eta$. When $\eta$ is absent, it means there is no similarity-based weighting for the loss calculated based on pseudo-labels. That means the pseudo-labels are used as actual ground-truths and the network is basically trained in a supervised fashion on both labeled images and those in pairs. The scenario with labeled images and no pairs, no $\lambda$, and no $\eta$ is equivalent to a fully supervised version of our method when it is only trained on the $1118$ labeled images.

The results of the ablation study indicate that our method performs its best when we include the additional unlabeled data produced using our pairing algorithm as well as using both $\lambda$ and $\eta$ to avoid initial noisy prediction of the network and control the contribution of the pseudo-labels in the training process. In addition, we observe that the supervised version of HRNetV2 performs better than the semi-supervised version of other backbones, and the additional data from pairs without being carefully controlled by $\eta$ and $\lambda$ would hurt its performance. That is because HRNet maintains multiple high-resolution features in parallel throughout its training and exchanges information between them via a multi-scale fusion. HRNetV2 specifically outputs four-resolution representations containing rich and precise spatial information; therefore, it outperforms other backbones in supervised training, but it can be potentially more sensitive to noise if it is not controlled by $\eta$ and $\lambda$.

\begin{figure}[]
    \centering
    \includegraphics[width=0.85\columnwidth]{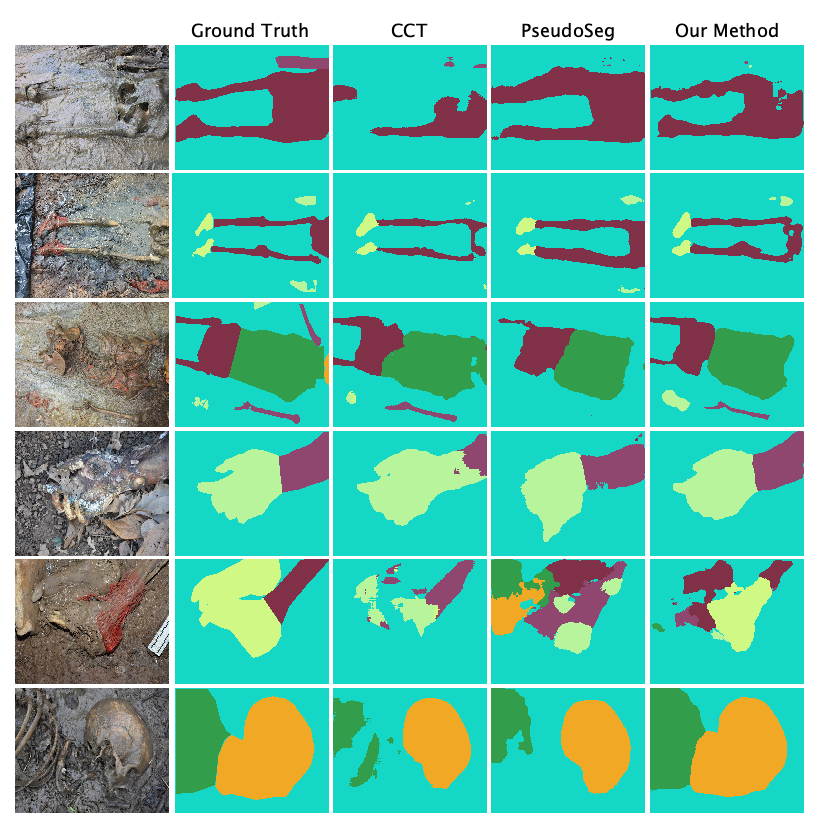}
    \caption{Examples of SLRNet and those from CCT and PseudoSeg on the human decomposition dataset when all methods are trained on $1118$ labeled images and $5906$ unlabeled images.}
    \label{fig:iccv_result}
\end{figure}

Furthermore, we qualitatively evaluate the performance of our method by providing the resulted segmentation for a few examples from using CCT, PseudoSeg, and SLRNet (examples shown in Figure \ref{fig:iccv_result}). Results indicate that our method captures classes and their annotations more accurately and is more conservative with less false positives and more false negatives (first and third row) compared to CCT and PseudoSeg. We believe the reason is because of how our ``reuse'' technique promotes regions that are shared between pairs amalgamating to more conservative masks.

\subsection{Limitations}
The key assumption of SLRNet is that the image collection contains unannotated images that are sufficiently similar to annotated images. Furthermore, these similar images contain annotations that are similarly located. Such instances tend to be common in collections where the same or similar object or subject occurs in multiple images as in our case of human decomposition and growing plants. Our approach may not be suitable for image collections that do not exhibit this property as illustrated in supplemental materials for VOC12 dataset. In this dataset a subject or an object is almost never photographed multiple times, so its annotation can not be re-used. Also, such datasets do not follow a specific photography protocol (common in many industry and research datasets) that may help partially align the annotations. In research and industry practice, the image collections very often contain repeated and similar images that are taken using a specific protocol. We, therefore, believe that the assumptions needed for SLRNet to perform on real datasets well are satisfied quite often. In these cases, SLRNet first identifies such opportunities where images are similar to one another and second adjusts for their potential annotation misalignments. 
% The key assumption of SLRNet is that the image collection contains unannotated images that are sufficiently similar to annotated images. Furthermore, these similar images contain annotations that are similarly located. Such instances tend to be common in collections where the same or similar object or subject occurs in multiple images as in our case of human decomposition and growing plants. Our approach may not be suitable for image collections that do not exhibit this property as illustrated in supplemental materials for VOC12 or Cityscapes datasets. In these datasets a subject or an object are almost never photographed multiple times, so its annotation can not be re-used. Also, these datasets do not follow a specific photography/video protocol (common in many industry and research datasets) that may help partially align the annotations. In research and industry practice, the image collections very often contain repeated and similar images that are taken using a specific protocol. We, therefore, believe that the assumptions needed for SLRNet to perform on real datasets well are satisfied quite often. In such cases, SLRNet first identifies such opportunities where images are similar to one another and second adjusts for their potential annotation misalignments. 

%% file: conclusion.tex
In this paper, we presented~\method, a simple semi-supervised technique for semantic segmentation of human decomposition data. The data at hand is a large image collection of decaying body parts with significant importance to the forensic community. Characteristically, the images in the data evolve through time as body parts decay and few labels are available due to the cost and scarcity of domain experts. Our main idea is to exploit image similarities and transfer labels from labeled to similar unlabeled images. To handle instances where the otherwise similar images do not contain the target in the same location, we structured a loss function to controls their level of impact on the network's learning. 
We tested our method on a subset of our human decomposition imagery with $1118$ labeled images and $5906$ unlabeled images. Results show that our method outperforms the state-of-the-art CCT and PseudoSeg methods in semantic segmentation task of human decomposition data, while being faster and less conceptually complex. %In future works, we would like to test our method on other datasets with similar characteristics with evolving content.

% Concerns:
% No GAN in related works
% No pose detection networks in related works
% BG ... results